# Active Wildfires Detection and Dynamic Escape Routes Planning for Humans through Information Fusion between Drones and Satellites

Chang Liu, *Member, IEEE,* Tamas Sziranyi, *Senior Member, IEEE*

*Abstract*— UAVs are playing an increasingly important role in the field of wilderness rescue by virtue of their flexibility. This paper proposes a fusion of UAV vision technology and satellite image analysis technology for active wildfires detection and road networks extraction of wildfire areas and real-time dynamic escape route planning for people in distress. Firstly, the fire source location and the segmentation of smoke and flames are targeted based on Sentinel 2 satellite imagery. Secondly, the road segmentation and the road condition assessment are performed by *D-linkNet* and NDVI values in the central area of the fire source by UAV. Finally, the dynamic optimal route planning for humans in real time is performed by the weighted *A\** algorithm in the road network with the dynamic fire spread model. Taking the Chongqing wildfire on August 24, 2022, as a case study, the results demonstrate that the dynamic escape route planning algorithm can provide an optimal real-time navigation path for humans in the presence of fire through the information fusion of UAVs and satellites.

## I. Introduction

Drone vision technology is being used in an increasing number of scenarios. In agriculture, drones equipped with high-resolution cameras or sensors can conduct aerial surveys of farmland to monitor the growth of crops, pests and diseases, and soil moisture conditions [1-3]. In the field of natural disaster rescue, drones are used to search, and rescue trapped people, monitor the situation in disaster areas, drop life-saving equipment and medical supplies to disaster areas, and provide first aid support [4-7]. In the field of logistics, drones can provide efficient courier and logistics services in urban and rural areas. They can quickly and accurately deliver packages from warehouses or sorting centers to their destinations, saving time and labor costs [8-9]. In the defense sector, swarms of drones can be used for missions such as target reconnaissance, intelligence gathering, border patrol, reconnaissance and surveillance [10-11]. And so on. For the application of drones in the field of natural disaster rescue, drones are a potentially efficient rescue tool when natural forest fires or man-made fires occur in the wild, and they have great flexibility and speed [12]. In addition to the time-sensitive drones, satellites can also provide us with more data on wildfires [13]. For example, the Sentinel 2 has 13 spectral bands, ranging from visible to near infrared to short-wave infrared, with spatial resolutions from 10 m to 60 m. Composed of two satellites, Sentinel 2A and Sentinel 2B, the temporal resolution of image acquisition is 10 days per satellite, with two complementary satellites and a revisit period of 5 days [14]. Usually, wildfires burn continuously for a period of time, days or even months, so the data provided by satellites is especially important at this time [15-17].

In this paper, we propose a fusion technique that cooperates the long-term data provided by satellites with the data collected by drones in the present moment in wildfires and finally provide dynamic optimal escape routes for humans in the active wildfire. First, based on the data provided by Sentinel 2 during the wildfire we can perform satellite image analysis, and based on the information provided by the different bands, we can roughly determine the location of the fire source and the direction of its spread. Secondly, the next role is played by the flexible drone fleet, they can quickly fly to the central area of the fire source for detection and rescue based on the location information from the previous step and carry out road network segmentation in the fire area through drone vision technology. Assessment of road conditions and navigation of the best dynamic escape routes is carried out through the road network maps acquired by the UAVs. The information fusion between the two makes wildfire detection and rescue efforts more efficient and accurate.

The main contributions and innovations of this work are as follows:

- Enabling information fusion between drones and satellites in wildfire rescue.
- Incorporating a dynamic fire spread model into the UAV-extracted road network map.
- Realizing the task of UAVs to plan dynamic escape paths in wildfire rescue in real time.

In the subsequent sections, Section 2 describe the work that has been done. In Section 3, the overall system architecture and the core algorithms for the main dynamic optimal escape route planning are presented. In the next section, we present some results based on the mountain wildfires that occurred in Chongqing on August 24, 2022. Finally, we conclude this work with a summary and outlook.

## II. Related Works

### A. Sentinel 2 Satellite Image Wildfire Analysis

Dedicated to supplying data for Copernicus services, Sentinel-2 carries a multispectral imager with a swath of 290 km. The imager provides a versatile set of 13 spectral bands spanning from the visible and near infrared to the shortwave infrared, featuring four spectral bands at 10 m, six bands at 20

Chang Liu is with the Department of Networked Systems and Services, Budapest University of Technology and Economics, Műegyetem rkp. 3, H-1111 Budapest, Hungary and Machine Perception Research Laboratory of Institute for Computer Science and Control (SZTAKI), H-1111 Budapest, Kende u. 13-17, Hungary (corresponding author to provide phone: +36-30-526-4122; e-mail: changliu@hit.bme.hu).

Tamas Sziranyi is with the Faculty of Transportation Engineering and Vehicle Engineering, Budapest University of Technology and Economics (BME-KJK), Műegyetem rkp. 3, H-1111 Budapest, Hungary and Machine Perception Research Laboratory of Institute for Computer Science and Control (SZTAKI), H-1111 Budapest, Kende u. 13-17, Hungary (e-mail: sziranyi.tamas@sztaki.hu).

m and three bands at 60 m spatial resolution. As indices primarily deal with combining various band reflectances, the Table 1 of 13 bands is given here for reference. The names of the Sentinel-2 bands are B01, B02, B03, B04, B05, B06, B07, B08, B8A, B09, B10, B11 and B12. The satellite data for the case used in this paper are images of a wildfire that occurred on August 24, 2022, in Jinyun Mountain, Beibei District, Chongqing, China, with different visualizations, from Sentinel 2. Calculations are performed from different bands of satellite images to visualize fire information [18]. Through the analysis and calculation of the following four images, we can well locate the location of the fire source and the direction of fire spread, which lays the foundation for the next tasks performed by the UAVs.

- RGB images: RGB three channels can get color image, the disadvantage of color image is that they cannot locate the fire source very well. The bands used are B02, B03 and B04.

- Active fire detection: Sentinel-2 identifies active fire points, offering valuable information for monitoring fire spots in any area. The bands used are B02, B03, B04, B11, and B12.

- Wildfire visualization: The bands used for fire visualization are B01, B02, B03, B04, B08, B8A, B11 and B12.

- Fire boundary image: The boundary of the affected wildfire area is important to understand and measure the impact of the event. It is important to highlight the boundary of affected areas in more contrast and detail. The bands used are B11 and B12 from Sentinel-2.

TABLE I. SPECTRAL BANDS FOR THE SENTINEL-2 SENSORS (S2A & S2B)

| Band Number | S2A & S2B | | |
|---|---|---|---|
| | S2A Bandwidth (nm) | S2B Bandwidth (nm) | Spatial resolution (m) |
| 1 | 20 | 20 | 60 |
| 2 | 65 | 65 | 10 |
| 3 | 35 | 35 | 10 |
| 4 | 30 | 31 | 10 |
| 5 | 14 | 15 | 20 |
| 6 | 14 | 13 | 20 |
| 7 | 19 | 19 | 20 |
| 8 | 105 | 104 | 10 |
| 8a | 21 | 21 | 20 |
| 9 | 19 | 20 | 60 |
| 10 | 29 | 29 | 60 |
| 11 | 90 | 94 | 20 |
| 12 | 174 | 184 | 20 |

MSI Instrument – Sentinel-2 MSI Technical Guide – Sentinel Online.

## B. Road Networks Extraction and Road Condition Assessment

In our previous work [20], we addressed the problem of segmenting roads and assessing road conditions using fusion images between drones and satellites. The road segmentation is based on *D-linkNet* [19] to extract the road network, and then the NDVI index values are computed using images captured from satellites or multispectral UAVs, and the segmented road network is assigned weights based on the NDVI index values and the evaluation of the road throughput, i.e., the road network map with different priorities is obtained. Figure 1 is a figure from our previous work [20], from which we can see that the roads with different road conditions are well separated based on the classification of NDVI index values and the segmentation of the road network. The road network diagram with weights in the lower right corner shows us that the green roads represent the roads with good road conditions and the white roads represent the roads with bad road conditions.

Figure 1. Road extraction and road condition evaluation by UAVs [20].

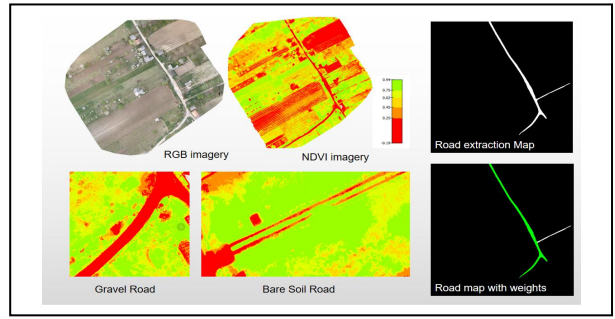

## III. METHODOLOGY

This section is the methodology and focuses on the main framework of the system proposed in this paper and how the core dynamic optimal escape route navigation algorithm works.

### A. Framework of the Proposed System

Figure 2 illustrates the working process of the entire system proposed in this paper. Firstly, the Sentinel-2 satellite can provide the system with multi-channel information, and based on the analysis of satellite images in different channels we can perform fire segmentation and smoke direction display. Next, based on the information provided in the previous step, we can lock the specific geographical location of the fire center, at this point the UAVs can go to the fire center area by virtue of its flexibility and convenience, as the UAVs are able to adjust the flight altitude during the flight process, so at this stage, the UAVs not only carry out the road network extraction task but also carry out the search and rescue of human beings. The search task of drones for human recognition through rescue body gestures has been implemented in our work [7]. Finally, based on the extraction and segmentation of the road network and the assessment of road conditions, the UAVs start to reconnaissance and perform dynamic escape route navigation tasks, as the fire is spreading dynamically, so the UAVs need to search for the best escape route for humans and vehicles in real time, to achieve the ultimate goal of timely and effective rescue.

Figure 2. Wildfire detection and dynamic escape route planning system.

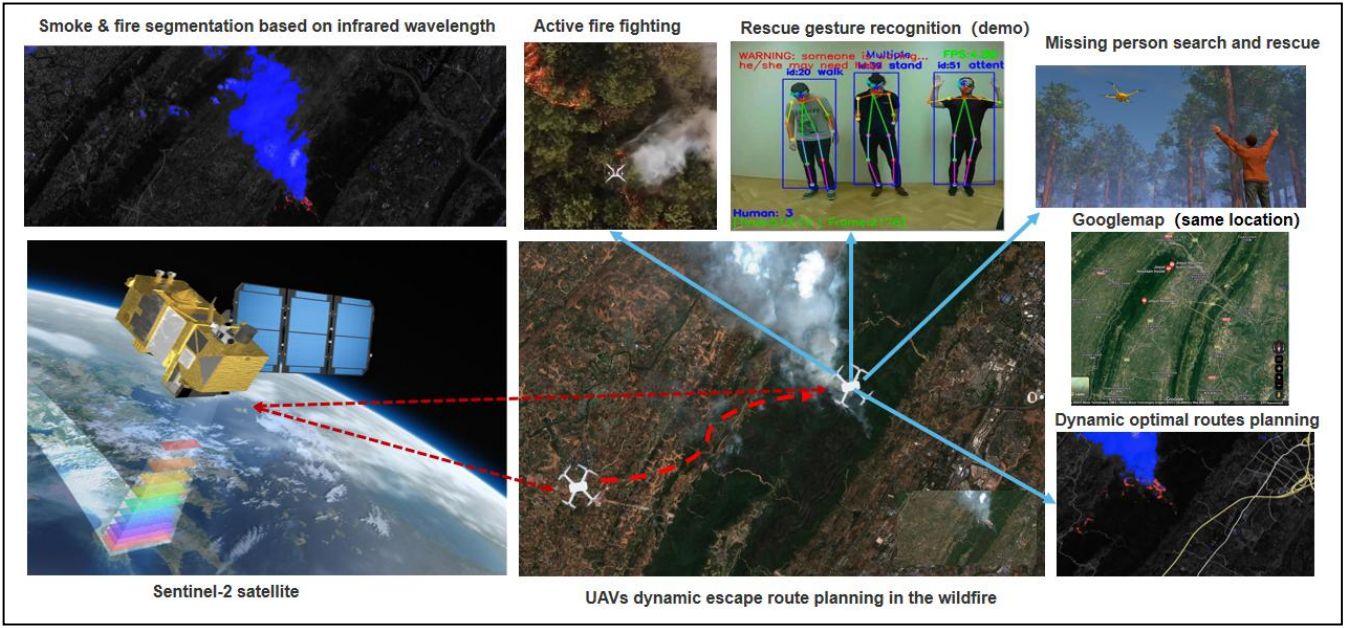

## B. Weighted A* Algorithm with Dynamic Fire Model

This work was run in Python on a Windows system ground station with with a GTX Titan GPU and an onboard UAV with Jetson Xavier GPU, importing the three necessary libraries, OpenCV, NumPy, and Random, for constructing dynamic fire model. For the road network images transmitted in from the UAVs, the model uses a spread probability variable to define the probability of fire spreading to neighboring cells, wind direction and wind speed variables to define wind direction and wind speed, number steps variable defines the number of time steps of the simulation, the fire $x$ and fire $y$ variables define the location of the fire source, and the fire radius variable defines the initial radius of the fire. Next, a loop is used to simulate the fire propagation for the given number of time steps. The new location of the fire source is calculated based on the wind direction and wind speed. The wind direction is randomly updated using a rotation matrix to generate a binary mask of neighboring cells around the fire source. Using the propagation mask, the fire is propagated to neighboring cells with a certain probability to increase the radius of the fire, and a red circle is drawn at the location of the fire source. The speed of fire propagation is depending on the the flammability of the surface material, what can be estimated from the last scanned images from the satellites. The above is the construction of the whole dynamic fire spread model.

Simultaneously with the fire spread model is the search for the dynamic best path, we have modified the $A*$ algorithm. $A*$ algorithm uses a combination of heuristic searching and searching based on the shortest path [21]. It is defined as the best-first algorithm because each cell in the configuration space is evaluated by the value:

$$f(n)=h(n)+g(n) \qquad (1)$$

where $g(n)$ represents the cost from the starting point to the current node; $h(n)$ represents the estimated cost from the current node to the ending point; n is the current node. $f(n)$ is the total cost of the node.

There are several well-known heuristic mathematical functions $h(n)$ that can be used, the most commonly used are Euclidean distance $h_E$, Manhattan distance $h_M$, or Diagonal distance $h_D$. In this work, we have chosen to use the $h_D$ to calculate the diagonal distance with the weighted modification [22]:

$$h_D(n) = d_1 * \max(dx, dy) + (d_2 - d_1) * \min(dx, dy) \qquad (2)$$

Where $(x_n, y_n)$ is the coordinate of the current node n; $(x_g, y_g)$ is the coordinate of the end node n; $dx$ is the absolute value of $x_n$ and $x_g$, $dy$ is the absolute value of $y_n$ and $y_g$. For green cell $d_1=1$ and $d_2=1.4$ (octile distance), white cell $d_1=100$ and $d_2=140$.

Unlike our previous work [22], the case of the $g(n)$ function in this task has changed. The original $g(n)$ function only needs to consider green pixels and white pixels, that is, roads in good condition and roads in bad condition, and the rest are all black pixels, which are not accessible to the search algorithm, that is, black pixels are considered as obstacles. In this work, because the dynamic fire model needs to be taken into account, it is possible for red pixels to appear on the roads on the original green and white pixels, where the red pixels refer to flames, because as the fire spreads, wildfires can block sections of the roads and bring threats to people trying their best to escape. Based on the above, we classified and discussed the situation of $g(n)$ functions in the algorithm, and roughly we divided them into 4 categories, as shown in Figure 3 to Figure 5.

Figure 3. Different assignment of *g(n)* cost value to different pixel points (Case 1 and Case 2).

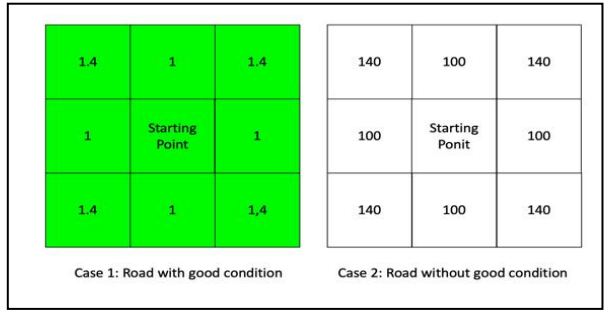

- Case 1 is when the starting pixel is located at a green pixel and surrounded by green pixels, which means that the user's starting position is in an environment where the surrounding roads are good and there is no fire threat, then the eight surrounding pixels are assigned a value of 1 or 1.4.
- Case 2 is when the starting pixel is in white and surrounded by white pixels, which means that the user's starting location is on a road that is not in good condition, but there is no fire threat from red pixels, so the eight neighboring pixels are assigned values 100 times higher than in case 1, with values of 100 and 140.

Figure 4. Different assignment of *g(n)* cost value to different pixel points (Case3).

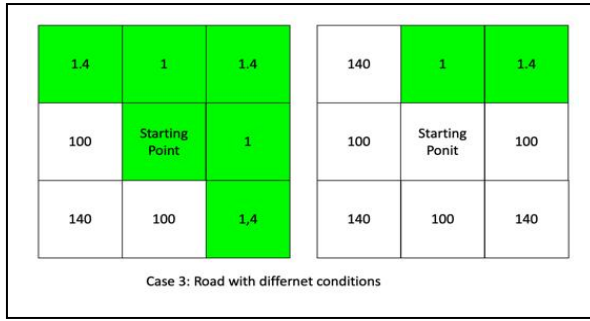

- Case 3 is where the user's starting point may be a white pixel with a not very good road condition or a green pixel with a good road condition, then the different value assignments come into play and the algorithm goes to find the path with the least cost, i.e. the algorithm tries to use pixels with a good road condition.
- Case 4 is the most complicated one. Based on the introduction of case 3, the algorithm is not allowed to visit red pixels when red wildfire pixels are encountered, and for realistic considerations, the algorithm should choose the path with good road conditions for humans to escape while avoiding wildfires.

Figure 5. Different assignment of *g(n)* cost value to different pixel points (Case4).

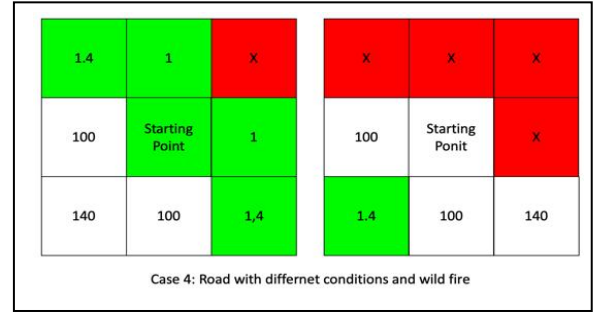

The above is the classification and introduction of the *h(n)* and *g(n)* functions, based on the modified algorithm, then the user gives a start and end point for each, the algorithm can well avoid the spread of dynamic fire, to find an optimal escape route in real time.

IV. EXPERIMENTAL RESULTS

This paper presents the results of the Jinyun Mountain wildfire in Chongqing China on 24 August 2022 as a case study. This section demonstrates the monitoring of active wildfires by satellite imagery and the segmentation and condition evaluation of roads in fire areas by drones, as well as providing dynamic and optimal escape routes in real time for humans in distress within seconds.

A. Active Wildfire Detection by Sentinel-2

Based on the section II of the related work, we can visualise the state of the active wildfire based on 13 channels of satellite data [18]. The first is the RGB three-channel result, from which we can see that the colour image does not clearly locate the source of the fire, which is covered by the upper layers of smoke. But based on the information from the other channels, we can clearly show the position of the fire center and the direction of the smoke, we can also visualise the wildfire and finally we can also get an image of the wildfire boundary, which can also help us to analyse the future trend of the fire.

Figure 6. Wildfire visualization results based on Sentinel 2 satellite imagery[18].

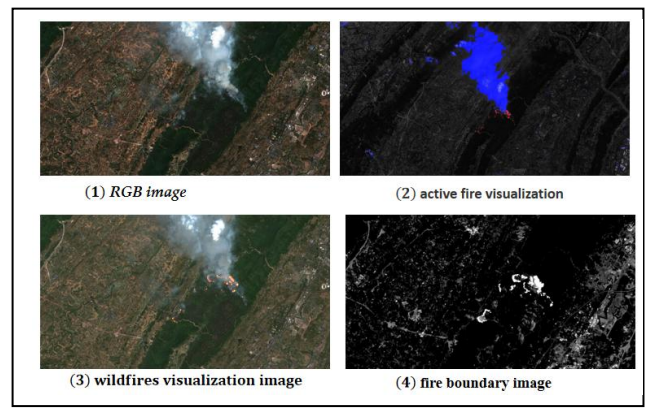

## B. Comparison Results of Dynamic Escape Route Planning

With the information about the wildfire from the Sentinel-2 satellite, we can well get the geographic location of the fire center of this wildfire disaster and the spread of the fire. Figure 7 shows the satellite image from the same geographic location with Google Maps image, we can see that the vicinity is wild mountain forest, and some villages also appear in the vicinity, we use a more complex road network map to verify that the subsequent dynamic path navigation algorithm is effective.

Figure 7. Road Extraction and Road Condition Evaluation Result in the Wildfire.

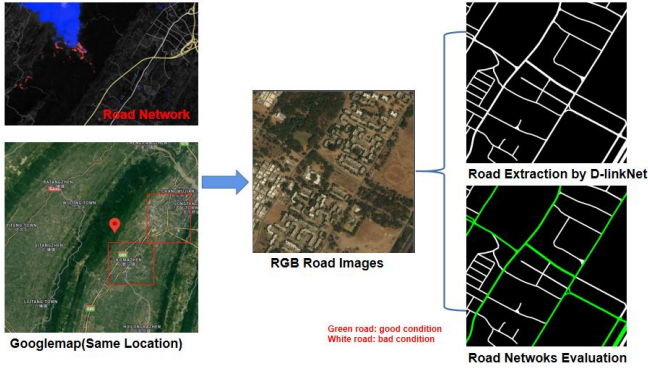

We demonstrate the dynamic fire spread and dynamic path planning on this extracted complex road network in Figure 7 to verify the effectiveness of the algorithm proposed in this work. Figures 8 to 10 show the comparative results of the three experimental groups. In each diagram, the green road represents the road in good condition, the white road represents the road in poor condition, the blue circle represents the user's starting location, the yellow circle represents the user's destination, the red area represents the spread of the fire over time, and the red line is the result of the final dynamic best path planning.

Figure 8. Dynamic optimal escape route planning result comparison 1.

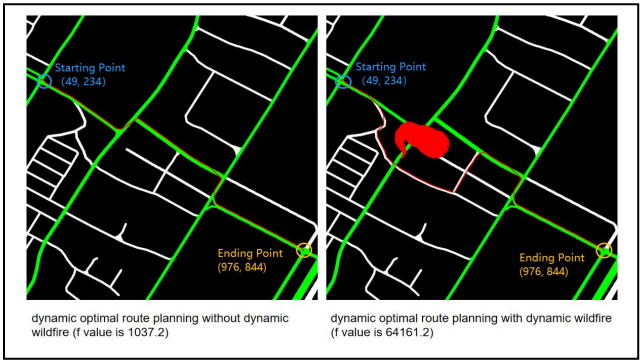

From the comparison chart in Figure 8 we can see that the spread of the fire affects the best path planning results of the UAV. When there is no dynamic fire model, the best path will select the route with good road conditions to provide the best route for the user, and the value of the $f(n)$ function obtained at this time is 1037.2. On the contrary, the same starting point and destination, when encountering a dynamic wildfire, the algorithm has to avoid the fire area so as to find the best navigation route, and we can see that the value of $f$ becomes 64161.2.

Figure 9. Dynamic optimal escape route planning result comparison 2.

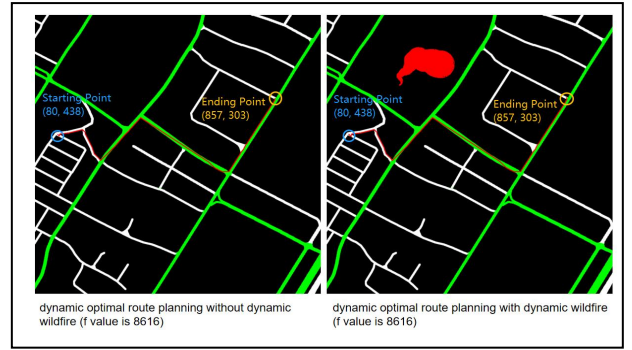

Figure 9 shows a situation where the fire spread has no effect on the UAV path planning. We can see from the comparison chart that the best paths planned by the red lines are the same for the same starting point, and the total cost value of the $f$ is also the same. This situation is because the location of the fire source at this time and the direction of the fire spread during the path planning will not affect the UAV navigation, so we can see that the two best paths are the same.

Figure 10. Dynamic optimal escape route planning result comparison 3.

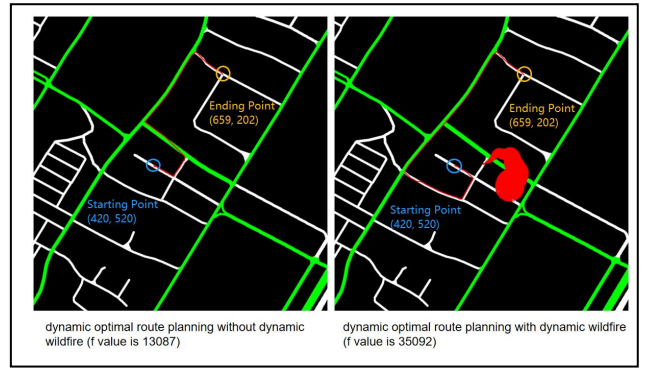

Figure 10 shows a more complicated case, where the same starting location, without dynamic fire spread, the best path is found by using roads in good condition as much as possible, resulting in a smaller total loss, with a value of $f(n)$ of 13087. In contrast, the plot on the right in Figure 10 shows well that the dynamic path planning algorithm avoids the center of the fire source and the direction of fire spread during that time, corresponding to a value of $f(n)$ function also increases a lot, with the value of $f$ reaching 35092.

## V. CONCLUSION

The main contribution of this paper is to use the information fusion between satellites and UAVs for wildfire detection and real-time dynamic rescue route planning. Taking the Jinyun Mountain wildfire that occurred in Chongqing, China on August 24, 2022 as an example, we fuse satellite image information and real-time UAV capture information to determine the location of the fire source and the tendency of the fire, use the flexibility of the UAV to extract the road segmentation and perform road condition assessment. Finally, in the case of dynamic fire spread, the UAV plans the best escape route in real time in the road network for the distressed human beings in distress to navigate rescue.

The results in this paper verify that the dynamic escape path planning algorithm is well suited to help humans avoid danger and provide an optimal route in the face of spreading fires. For future work, we will include a comparison of the efficiency between different path search algorithms, and an evaluation of different state-of-art road extraction network algorithms.

ACKNOWLEDGMENT

The work is carried out at Institute for Computer Science and Control (SZTAKI), Hungary, and the authors would like to thank their student Morui Zhu for the Chongqing wildfire Sentinel-2 dataset. The research work was supported by project no. 2022-2.1.1-NL-2022-00012 National Laboratory of Cooperative Technologies is funded by the Ministry of Culture and Innovation through the National Research, Development and Innovation Fund, under the funding of the 2022-2.1.1-NL Establishment and Complex Development of National Laboratories call, by the European Union within the framework of the National Laboratory for Autonomous Systems (RRF-2.3.1‐21-2022‐00002) programs, and by the TKP2021-NVA-01 project.

REFERENCES

[1] Zhang, H., Wang, L., Tian, T. and Yin, J., 2021. A review of unmanned aerial vehicle low-altitude remote sensing (UAV-LARS) use in agricultural monitoring in China. Remote Sensing, 13(6), p.1221.

[2] Boursianis, A.D., Papadopoulou, M.S., Diamantoulakis, P., Liopa-Tsakalidi, A., Barouchas, P., Salahas, G., Karagiannidis, G., Wan, S. and Goudos, S.K., 2022. Internet of things (IoT) and agricultural unmanned aerial vehicles (UAVs) in smart farming: a comprehensive review. Internet of Things, 18, p.100187.

[3] Reinecke, M. and Prinsloo, T., 2017, July. The influence of drone monitoring on crop health and harvest size. In 2017 1st International conference on next generation computing applications (NextComp) (pp. 5-10). IEEE.

[4] Dong, J., Ota, K. and Dong, M., 2021. UAV-based real-time survivor detection system in post-disaster search and rescue operations. IEEE Journal on Miniaturization for Air and Space Systems, 2(4), pp.209-219.

[5] Mishra, B., Garg, D., Narang, P. and Mishra, V., 2020. Drone-surveillance for search and rescue in natural disaster. Computer Communications, 156, pp.1-10.

[6] Pathan, A.I., Kulkarni, M.A.P., Gaikwad, M.N.L., Powar, M.P.M. and Surve, A.R., 2020. An IoT and AI based Flood Monitoring and Rescue System. Int. J. Eng. Tech. Res, 9(9), pp.564-567.

[7] Liu, C. and Szirányi, T., 2021, April. Gesture Recognition for UAV-based Rescue Operation based on Deep Learning. In Improve (pp. 180-187).

[8] Perera, S., Dawande, M., Janakiraman, G. and Mookerjee, V., 2020. Retail deliveries by drones: how will logistics networks change?. Production and Operations Management, 29(9), pp.2019-2034.

[9] Roca-Riu, M. and Menendez, M., 2019. Logistic deliveries with drones: State of the art of practice and research. In 19th Swiss Transport Research Conference (STRC 2019). STRC.

[10] Lehto, M. and Hutchinson, B., 2020, March. Mini-drones swarms and their potential in conflict situations. In 15th international conference on cyber warfare and security (pp. 326-334).

[11] Lehto, M. and Hutchinson, W., 2021. Mini-Drone Swarms. Journal of Information Warfare, 20(1), pp.33-49.

[12] Daud, S.M.S.M., Yusof, M.Y.P.M., Heo, C.C., Khoo, L.S., Singh, M.K.C., Mahmood, M.S. and Nawawi, H., 2022. Applications of drone in disaster management: A scoping review. Science & Justice, 62(1), pp.30-42.

[13] Filipponi, F., 2018, March. BAIS2: Burned area index for Sentinel-2. In Proceedings (Vol. 2, No. 7, p. 364). MDPI.

[14] Drusch, M., Del Bello, U., Carlier, S., Colin, O., Fernandez, V., Gascon, F., Hoersch, B., Isola, C., Laberinti, P., Martimort, P. and Meygret, A., 2012. Sentinel-2: ESA's optical high-resolution mission for GMES operational services. Remote sensing of Environment, 120, pp.25-36.

[15] Coen, J.L. and Schroeder, W., 2013. Use of spatially refined satellite remote sensing fire detection data to initialize and evaluate coupled weather-wildfire growth model simulations. Geophysical Research Letters, 40(20), pp.5536-5541.

[16] Edwards, D.P., Emmons, L.K., Hauglustaine, D.A., Chu, D.A., Gille, J.C., Kaufman, Y.J., Pétron, G., Yurganov, L.N., Giglio, L., Deeter, M.N. and Yudin, V., 2004. Observations of carbon monoxide and aerosols from the Terra satellite: Northern Hemisphere variability. Journal of Geophysical Research: Atmospheres, 109(D24).

[17] Zhang, X., Kondragunta, S., Ram, J., Schmidt, C. and Huang, H.C., 2012. Near-real-time global biomass burning emissions product from geostationary satellite constellation. Journal of Geophysical Research: Atmospheres, 117(D14).

[18] Sinergise, Sentinel-Hub by. "Home." Sentinel Hub Custom Scripts, custom-scripts.sentinel-hub.com/. Accessed 15 May 2023.

[19] Zhou, L., Zhang, C. and Wu, M., 2018. D-LinkNet: LinkNet with pretrained encoder and dilated convolution for high resolution satellite imagery road extraction. In Proceedings of the IEEE Conference on Computer Vision and Pattern Recognition Workshops (pp. 182-186).

[20] Liu, C. and Szirányi, T., 2022. Road Condition Detection and Emergency Rescue Recognition Using On-Board UAV in the Wildness. Remote Sensing, 14(17), p.4355.

[21] Duchoň, F., Babinec, A., Kajan, M., Beňo, P., Florek, M., Fico, T. and Jurišica, L., 2014. Path planning with modified a star algorithm for a mobile robot. Procedia engineering, 96, pp.59-69.

[22] Liu, C. and Szirányi, T., 2022, June. UAV Path Planning based on Road Extraction. In IMPROVE (pp. 202-210).